\pdfoutput=1

\documentclass[11pt]{article}

\usepackage{acl}

\usepackage{times}
\usepackage{latexsym}

\usepackage[T1]{fontenc}

\usepackage[utf8]{inputenc}

\usepackage{microtype}

\usepackage{amsfonts}
\usepackage[mathscr]{eucal}
\usepackage[vlined,ruled,linesnumbered]{algorithm2e}
\usepackage{graphicx}
\usepackage{mathtools}
\DeclarePairedDelimiter{\norm}{\lVert}{\rVert} 
\usepackage{booktabs}
\usepackage{amssymb}
\usepackage{diagbox}
\usepackage{subcaption}

\title{Training Mixed-Domain Translation Models via Federated Learning}

\author{Peyman Passban\Thanks{ Equal contribution.}\hspace{7mm} Tanya Roosta$^*$\hspace{7mm}\\
\textbf{Rahul Gupta\hspace{7mm} Ankit Chadha\hspace{7mm} Clement Chung}\\
  Amazon\\
  \texttt{\{peymp,troosta,gupra,ankitrc,chungcle\}@amazon.com} \\}

\begin{document}
\maketitle
\begin{abstract}
Training mixed-domain translation models is a complex task that demands tailored architectures and costly data preparation techniques. In this work, we leverage federated learning (FL) in order to tackle the problem. Our investigation demonstrates that with slight modifications in the training process, neural machine translation (NMT) engines can be easily adapted when an FL-based aggregation is applied to fuse different domains. Experimental results also show that engines built via FL are able to perform on par with state-of-the-art baselines that rely on centralized training techniques. 

We evaluate our hypothesis in the presence of five datasets with different sizes, from different domains, to translate from German into English and  discuss  how  FL  and  NMT  can mutually benefit from each other. In addition to providing benchmarking results on the union of FL and NMT, we also propose a novel technique to dynamically control the communication bandwidth by selecting impactful parameters during FL updates. This is a significant achievement considering the large size of NMT engines that need to be exchanged between FL parties.
\end{abstract}

\section{Introduction}
Federated learning (FL) is a rapidly growing field in the machine learning community. The reason for this popularity is because of its \textit{decentralized} and \textit{private} nature. Model training in FL is distributed over multiple nodes where each node could be an independent piece of hardware with its own isolated data. This unique feature enables building high-quality models that benefit from external resources without requiring access to them.

Although FL is a relatively new field \citep{mcmahan2017communication}, it has drawn researchers' attention and the community has witnessed a rapid growth. Fields such as computer vision have adapted quickly to the FL framework \citep{geyer2017differentially,lin2017deep,hardy2019md,geiping2020inverting,Ren2021GRNNGR}, but others, such as natural language processing (NLP), have not been as quick and only recently have begun to explore \citep{wu2020fedmed}. 

We believe that the reason for this slow(er) integration of NLP and FL is because representation learning in NLP is complicated and downstream tasks require large and data-hungry models. These requirements are heavy-handed for any FL model and slow down the unification. However, real-world NLP problems necessitate the use of distributed solutions with privacy-preserving characteristics \citep{9084352}. Our work is an effort towards leveraging FL-based solutions in NLP. 

In this paper, we focus on NMT to combine it with FL. A review of the NLP literature shows that almost all recent groundbreaking architectures have been first proposed, or at least evaluated, for translation \cite{bahdanau2014neural,sutskever2014sequence,gehring2017convolutional,vaswani2017attention}. This is an indication of the intricacy of the NMT task. Therefore, it is fair to claim that any FL technique that is capable of training high-quality NMT models could also be considered as a trustworthy alternative for other NLP tasks. Thus, NMT could be a strong candidate for FL benchmarking purposes. 

NMT also has other unique features that can directly benefit FL. One key factor in a fair simulation of FL is to mimic data heterogeneity (also referred to as non-IIDness) in experimental environments \citep{kim2020on}. Different sampling techniques have been proposed to model such data distribution processes \citep{ji2020dynamic,li2021federated,wang2021device}, however there is no guarantee that what we simulate is what we encounter in real-world applications. NMT, to some extent, solves this problem since parallel datasets by nature have such heterogeneity. There are only a few NLP tasks that can provide as large and diverse training corpora as NMT. For some language pairs, there exist multiple datasets with hundreds of thousands or even millions of parallel sentences.\footnote{\url{https://opus.nlpl.eu/}} This means, we should have enough data for each FL node. Moreover, each node can naturally pick one dataset/language, so we do not have to artificially distribute data. NMT, together with offering rich training data, also has a bi-lingual setting which is a compelling testbed for FL. Coping with the complexity of not only different domains, but also different languages, at the same time in a distributed platform is worth investigating. 

While NMT offers a natural testbed for FL, we argue that the task itself benefits from the offerings of FL. Training a mixed-domain translation model is a challenging task. 
We show that the aggregation phase of FL can greatly help with this challenge, as it efficiently fuses information from different domains. From the privacy perspective, NMT can also benefit from FL. In fact, the necessity of having a private training pipeline is relatively understudied in NMT. It is mostly assumed that all training datasets are available in a homogeneous and compatible format through a central repository, which is usually not attainable in real world. All these reasons make NMT a compelling case to study in the context of FL.

\subsection{Research Scope}
The goal of this paper is to provide preliminary results on the combination of FL and NMT, as opposed to running a comprehensive FL research or comparing different algorithms. We are also interested in studying the feasibility of training complex and deep NMT models in \textit{decentralized} and \textit{private} settings. Besides providing a set of benchmarking results, this paper's other two contributions can be summarized as follows:
\begin{itemize}
    \item We show that FL aggregation techniques are reasonable alternatives to fuse information from multiple domains, thus FL-based training could be considered as an approach to build mixed-domain NMT engines.
    
    \item We show that large NMT models are hard to distribute within the FL network. Therefore, we propose a novel and cost-efficient solution to reduce the communication bandwidth.
\end{itemize}

\section{Federated Learning}\label{background}
FL is an approach to train models in a distributed fashion where nodes do not (and are not allowed to) access each other's data \citep{yang2019federated,9084352}. Any node by itself is not powerful enough to deliver high-quality services due to the small size of its local data. It can perform well on in-domain instances, but it might fail to respond to requests from other domains. FL establishes a communication methodology and a platform that allows participating nodes to exchange parameters (but not data) to help boost each other's performance. 

In this work, we follow the \textit{cross-silo} FL setting as outlined in \citet{li2019survey}. Algorithm \ref{algo-1} summarizes the entire model training pipeline and explains what we mean by being cross-silo.
\begin{algorithm}[h]
\DontPrintSemicolon
\SetArgSty{textnormal}
\SetKw{KwBy}{by}
 \For{$r\gets0$ \KwTo \textit{rounds} \KwBy $1$}{
    updates = {\fontfamily{pcr}\selectfont Pull(}\hspace{0.5mm}$\mathcal{C}${\fontfamily{pcr}\selectfont)}\;
    \For{ \textit{upd} \text{\textbf{in}} \textit{updates}}{
    $\mathcal{S}$ =  \textit{aggregate}\hspace{0.5mm}($\mathcal{S}$,\textit{upd})\;
    }
    $\mathcal{C}$ = {\fontfamily{pcr}\selectfont Push(}\hspace{0.5mm}$\mathcal{S}${\fontfamily{pcr}\selectfont)}\;
    }
\caption{\label{algo-1}Cross-Silo FL}
\end{algorithm}

In this setup, there is a central node $\mathcal{S}$ that orchestrates training. In each round $r$, the server pulls local updates (i.e. a set of parameters) from different nodes (also known as clients) and updates the parameters of the central model. $\mathcal{C} = \{c_1, ..., c_K\}$ indicates the set of $K$ clients. Once all information is aggregated, parameters of the central model are pushed back to clients so that they can also benefit from global/community knowledge. 

One key factor in FL is \textit{communication}, which is defined by the {\fontfamily{pcr}\selectfont Pull} and {\fontfamily{pcr}\selectfont Push} steps in this algorithm. Due to the distributed nature of FL, nodes need to connect and exchange information and this needs to be carried out in an efficient fashion. The communication cost becomes even more critical when exchanging large models, such as in NMT. In the next sections, we discuss how communication directly affect the feasibility and performance of any FL setting, and how we improve it by our \textit{dynamic pulling} technique.

Algorithm \ref{algo-1} only shows the computation that occurs on the server side. It should be noted that each client is an independent silo that updates its internal model with local data. This algorithm only illustrates the main skeleton of the cross-silo setting and does not entail all the details of each step. 

In our experiments, we use the well-known {\fontfamily{pcr}\selectfont FedAVG} algorithm \citep{mcmahan2017communication} for \textit{aggregation}. Therefor, Line 4 can be formulated as in Equation \ref{eq1}: 
\begin{equation}\label{eq1}
    w_{r} \leftarrow \sum_{k=1}^{K} \frac{n_k}{n} w^k_r
\end{equation}
where $w_r$ is the set of all parameters of the central model in the \textit{r}-\textit{th} round, $n_k$ is the number of data points in the \textit{k}-th client's dataset, and $n$ is the total number of all training samples. There exist multiple extensions to {\fontfamily{pcr}\selectfont FedAVG}, but since it is a widely-acceptable baseline for benchmarking purposes we also use it in our experiments. This choice allows us to minimize the impact of different factors introduced by other FL algorithms and only focus on the relation between NMT and FL and their mutual impact on each other. 

\section{Federated Learning for NLP}
There are several models in the field that have been proposed to leverage FL for NLP. \citet{hartmann2019federated} studied whether they could improve the ranking of suggestions in the Firefox URL bar and train a model on user interactions without violating user privacy. They incorporated feedback received from different clients using {\fontfamily{pcr}\selectfont FedAVG} which resulted in significant improvements. 
\citet{ji2019learning} suggested that the simple averaging strategy used in {\fontfamily{pcr}\selectfont FedAVG} might not be sufficient enough, so they improved the aggregation phase by incorporating the significance of each client by using an attention mask to weigh clients. 
\citet{chen2019federated} focused on language modelling and addressed the problem of out-of-vocabulary entries when working with different clients.

\citet{bui2019federated} investigated the effect of FL in training better and more personalized user/data representations. Their results show that when aggregating information via FL, the model quality increases significantly; at the same time, the training pipeline is distributed and private. We also make a similar observation in our experiments, though not at the representation level, but in terms of the final translation quality. 

\citet{ge2020fedner} proposed a named-entity recognition (NER) model that is trained with FL to work on medical data. Their results demonstrate that not only FL preserves privacy, but also outperforms models trained in a centralized fashion. Apart from NER, models with similar concerns have been proposed for mobile keyword prediction \citep{Hard2018FederatedLF}, keyword spotting \citep{leroy2019federated}, and next word prediction \citep{stremmel2020pretraining}. 

\citet{liu2020federated} utilized FL to pre-train and/or fine-tune BERT models \citep{devlin2018bert}. From a research standpoint, it is worthwhile to understand if it is even feasible to handle such deep models in an FL framework, and whether a simple averaging-based aggregation is enough. They attempted to address these questions and provided supporting results. In that sense, our work is similar to theirs as we also work with complex models. In addition, we also discuss the bandwidth problem to facilitate exchanging large sets of parameters.

\subsection{Domain Adaptation}\label{da}
Domain adaptation covers a wide range of problems from adjusting a model to work in a new domain/genre \citep{chu-wang-2018-survey} to fine-tuning for noisier conditions \citep{passban2020revisiting}, or even transferring a model to a different environment for a different task \citep{zhu2020transfer}. Domain adaptation has recently attracted more attention due to advances introduced by models such as ELMO \citep{Peters:2018} and BERT \citep{devlin2018bert}. These models provide general-purpose representations which are easily adaptable to other tasks. In these models, \textit{all} network parameters are fine-tuned during adaptation, which might not be necessary. \citet{houlsby2019parameter}, \citet{pfeiffer2020adapterhub}, and \citet{ruckle2020adapterdrop} proposed a new set of architectures, known as Adapters, to tackle this problem. Adapters are low-cost plug-ins that are mounted on pre-trained models, so when adapting the model only these small sets of parameters are updated. 

\citet{bapna2019simple} proposed an NMT variation of Adapters. In their model, a dedicated component is added inside each layer that is responsible for transitioning in-between domains. However, all these solutions perform in centralized settings. \citet{Roosta2021CommunicationEfficientFL} studied this problem in the context of FL and showed that Adapters might not be aligned well with the distributed nature of FL. As they reported, Adapters seem to be suitable to connect two domains but when exposed to several domains in FL, they diverge too much from their main distribution, such that using them in the body of clients drastically downgrades performance. 

To address the problem, they introduced additional and dedicated layers (as opposed to intra-layer modules in Adapters), called Controllers, that are designed to be exchanged between the server and clients. These new layers are placed in-between client model's original layers and deal with external information sent/received to/from the server. Since they only exchange Controllers, they are able to reduce the communication bandwidth. This work is the closest to ours so we use it as our main baseline. 

The model proposed by \citet{Roosta2021CommunicationEfficientFL} suffers from two issues. They randomly introduce new layers (Controllers) but there is no mechanism defined to determine the number those layers, i.e. it is not investigated that how many Controllers are required under different conditions. It is also not clear where these layers need to be placed, and it is only discovered through experimental explorations. On the contrary, we introduce a simple yet effective heuristic to select a subset of impactful parameters during communication. In our solution, we do not need to deal with the aforementioned issues.

\section{Low-Cost Domain Adaptation in FL}\label{method}
In our FL setting, we initialize each client with a high-quality but generic NMT engine. Clients use local data to fine-tune their internal model and the combination of a pre-loaded model with local data should lead to better quality. Clients also connect with the server regularly to transfer local knowledge and contribute to the aggregation phase. In such a process, domain adaptation happens naturally. Inspired by \citet{Roosta2021CommunicationEfficientFL}, we implemented this idea and observed a substantial boost in our translation engines. Not only are we able to deliver better results but also we train NMT models in a distributed and private fashion. However, we noticed that \textit{communication} could be quite costly. 

\begin{figure}[htp]
    \centering
    \includegraphics[width=0.37\textwidth]{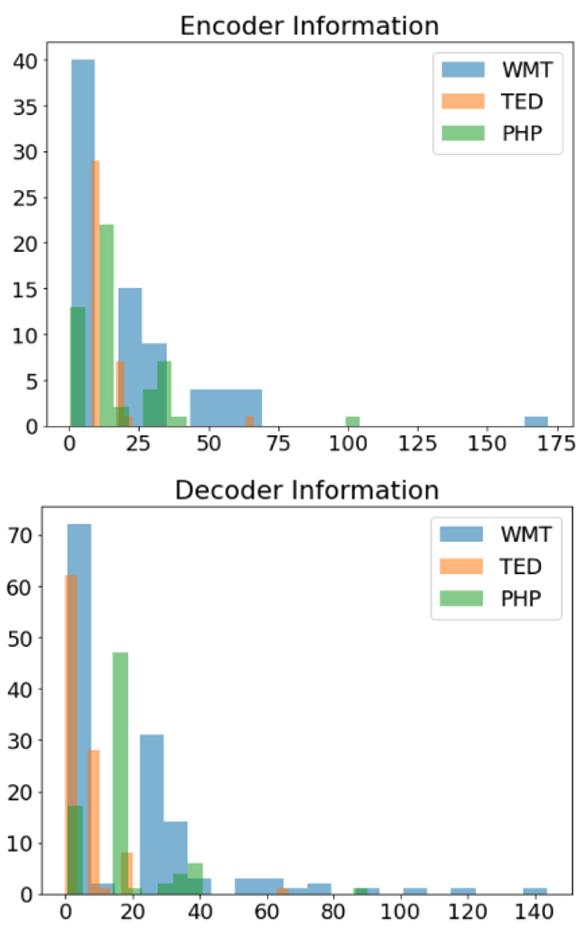}
    \caption{The histograms of the norm values of the difference tensors computed for the $110$K and $120$K checkpoints. The \textit{x} and \textit{y} axes show the \textit{norm} and the \textit{number of tensors}, respectively. Information related to encoder tensors is visualized in the upper half and the bottom sub-figure consists of decoder tensors' information. The first blue bar of the encoder sub-figure indicates that around $40$ tensors in the WMT encoder only changed slightly from the checkpoint $110$K to $120$K as their norm is in the range [0,5], whereas the last bar on the other end of the same sub-figure shows that around $2$ tensors in the 
    WMT model changed drastically as the norm of their difference tensors is close to $175$.}
    \label{fig:diff}
\end{figure}

In the default configuration, for every {\fontfamily{pcr}\selectfont Pull} (in Algorithm \ref{algo-1}) a large NMT engine has to be exchanged, which might not be necessary. In order to clarify why, we designed an experiment whose information is illustrated in Figure \ref{fig:diff}. In this experiment, we pick three of our engines and train them for $120$K steps within the FL pipeline. For each model, we compare tensors from the $120$K checkpoint to their $110$K peers and measure how much they changed in-between these two checkpoints.  More specifically, for a given tensor $w$, we compute the pair-wise difference between values from the two checkpoints ($w_{d} = w_{120K} - w_{110K}$), then compute the absolute-value norm of the difference tensor ($\norm{w_{d}}$). The norm value indicates the shift of each tensor during FL rounds. Figure \ref{fig:diff} provides a histograms of norm values that belong to different tensors from the encoders and decoders of our translation engines (for more information about the engines and datasets see Section \ref{exp}). 

Results from Figure \ref{fig:diff}, together with our other observations, show that a small subset of tensors diverge substantially (mostly shown in the right half of the figures), but for the rest, there is a heavy concentration around small norms. More interestingly, we realized that this is a pattern that consistently occurs from one round to the next in our FL experiments, namely each tensor either belongs to a set of highly-fluctuating parameters or it only changes marginally and lies in the less active set. The fluctuation threshold can change but tensors almost always stay in their respective clusters between rounds. We used this finding as a basis of our design to improve bandwidth consumption, such that we decided to focus on either highly fluctuating tensors or those in the other cluster and only {\fontfamily{pcr}\selectfont Pull} one type of tensors during communication. The strategy is simple but has led to promising results.

\begin{figure}[htp]
    \centering
    \includegraphics[width=0.4\textwidth]{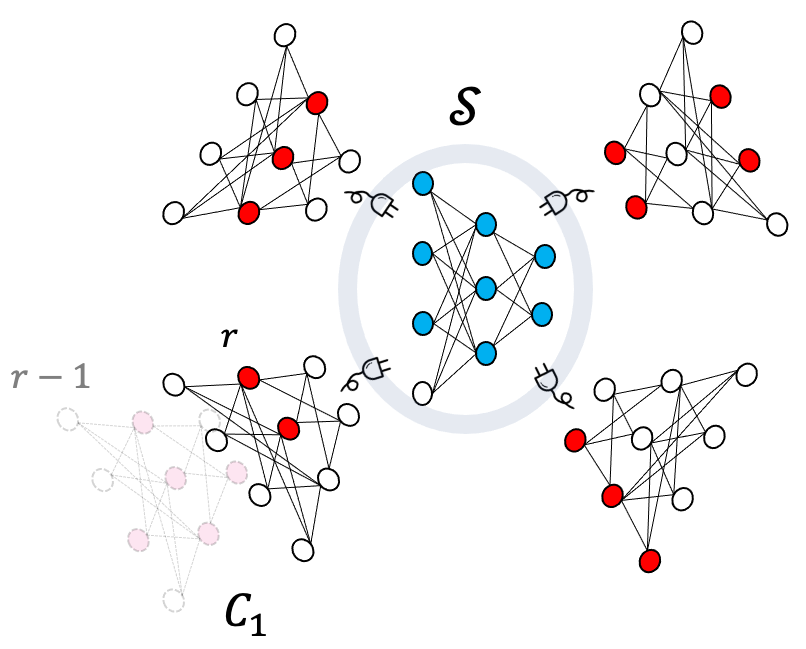}
    \caption{DP-based communication in our FL setting.}
    \label{fig:DP}
\end{figure}

More formally, a variation of the aforementioned idea can be simply formulated as shown in Equation \ref{eq:dp-gt}:
\begin{equation} \label{eq:dp-gt}
\begin{split}
DP_{g}^{c} = \{w_r^t; \norm{w_r^t - w_{r-1}^t} \geq \theta \} \\
\end{split}
\end{equation}
$DP_{g}^{c}$ in the \textit{r}-\textit{th} round consists of all tensors ($w^t_r$) that deviate from their previous values in the previous round by $\theta$. \textit{DP} stands for \textit{Dynamic Pulling} and \textit{g} indicates that the difference norm of candidate tensors should be \textit{greater than or equal} to the threshold. \textit{DP} is exclusively computed for each client which is specified with the \textit{c} superscript. Moreover, the \textit{DP} sets for decoders and encoders are calculated separately as they vary at different scales. Based on Equation \ref{eq:dp-gt}, we do not need to {\fontfamily{pcr}\selectfont Pull} all tensors but each client decides what to share with $\mathcal{S}$ (only highly-fluctuating tensors in this case). Figure \ref{fig:DP} visualizes this concept.

As shown in the figure, $C_1$ computes the difference between tensors from rounds $r$ and $r-1$ and decides to only share two tensors (vertices in diagrams) that have the highest norms. In this scenario, the communication cost between $C_1$ and $\mathcal{S}$ is approximately reduced by $78$\% for the pulling phase as only $2$ out of $9$ tensors ($22$\%) are transferred. The exact percentage of the bandwidth saved in this communication protocol directly depends on the client's architecture and $\theta$, but the figure uses an imaginary scenario to explain the reduction mechanism.

The intuition behind $DP_{g}^{c}$ is that only highly-fluctuating tensors should be involved in the communication process. It assumes that the main \textit{adaptation} (or \textit{out-domain}) knowledge lies in those tensors and what each client needs to learn about its external world is only communicated through such active tensors. Therefore, clients only need to exchange them with the server. This is an assumption that might either result in effective communication or conversely hurt the client. Because, if by any chance \textit{local} (or \textit{in-domain}) knowledge is stored in such tensors, $DP_{g}^{c}$ manipulates the most important parameters and overwrites them with external domains' information. In other words, there is a possibility that the reason for observing high fluctuations in active tensors is not because they carry the community knowledge but because they are responsible to learn local data, so they have to vary frequently to adjust and learn local data.

If this second assumption is correct, modifying the active set can easily delay the convergence of local models, and thus deteriorate their quality. Due to this concern, we propose another alternative, $DP_{l}^{c}$, which relies on less active tensors for communications. The pulling condition for $DP_{l}^{c}$ is as in Equation \ref{eq:dp-lt}:
\begin{equation} \label{eq:dp-lt}
\norm{w_r^t - w_{r-1}^t} < \theta
\end{equation}
which means, unlike the previous case, highly-active tensors are protected from external updates and only modified using local data. The less active tensors are assumed to be the representatives of external domains, so they are shared with the server and co-trained with other tensors. Both types of tensors contribute to the local training process, but this time the less active group is responsible for bringing the community knowledge to the client. In the next section, we compare $DP_{l}^{c}$ and $DP_{g}^{c}$ and show which strategy is more impactful.  


\section{Experimental Study}\label{exp}
\begin{table*}[h]
\begin{center}
\begin{tabular}{ l  l l  l l  l}
\toprule
\textbf{Corpus} & \multicolumn{2}{c}{\textbf{Source (De)}} & \multicolumn{2}{c}{\textbf{Target (En)}} & \textbf{Sentences}\\
& Words & Tokens & Words & Tokens & \\\midrule
 \textit{WMT} & 119,920,225 & 1,529,872  & 126,731,132 & 691,150 & 4,468,841\\ 
 \textit{OS} & 33,502,036 & 339,627 &  37,373,751 & 174,670 & 4,500,000\\
 \textit{TED} & 2,710,904 & 99,221 & 2,861,006 & 45,364 & 143,837\\
 \textit{PHP} & 322,546 & 13,001 & 318,788 & 8,521 & 39,708\\
 \textit{UB} & 96,355 & 13,515 & 90,839 & 9,234 & 13,246\\\bottomrule
\end{tabular}
\caption{\label{tab:stat} Statistics of the training datasets. The second and third columns show the number of all words and unique tokens for the source and target languages, respectively, and the last column is the number of parallel sentences. \textit{OS} is a large collection, so we randomly selected a subset of it for our experiments. As the statistics show, we have different sets with different sizes from different domains which helps us have a fair and realistic simulation.}
\end{center}
\end{table*}

\subsection{Hyper-Parameters and Datasets}
Since our main baseline is the model proposed in \citet{Roosta2021CommunicationEfficientFL}, we follow their setting in the interest of fair comparisons. For our  translation engines, we use Transformers with their \textit{base} configuration \citep{vaswani2017attention}. Encoders and decoders have six layers (each), attention modules have eight heads, word embeddings and internal projection layers are $512$-dimensional vectors, and the inner layer in the position-wise feed-forward module has $2024$ dimensions. All other hyper-parameters/variables such as the training algorithm and scheduling are the same as the \textit{base} configuration unless they are clearly indicated in the paper. We used four NVIDIA V100 GPUs for all experiments.   

Similar to \citet{Roosta2021CommunicationEfficientFL}, we work on the the German$\rightarrow$English direction. To train/test the models, we use five datasets of \textit{WMT},\footnote{\url{http://statmt.org/wmt14/translation-task.html}} \textit{OpenSubtitle} or \textit{OS} \citep{lison-tiedemann-2016-opensubtitles2016}, \textit{PHP}, \textit{Ubuntu} or \textit{UB}, and \textit{TED} \citep{TIEDEMANN12.463} where all corpora are \textit{normalized} and \textit{tokenized} with the scripts provided by Moses.\footnote{\url{https://github.com/moses-smt/mosesdecoder}} Table \ref{tab:stat} provides the statistics of the training datasets.

For the test and development sets of the \textit{WMT} model, we use \textit{newstest-14} and \textit{newstest-13}, respectively. For others, we randomly select 4,000 sentences: 2,000 for the test and 2,000 for the development sets.\footnote{The same sets used in \citet{Roosta2021CommunicationEfficientFL}} We also pre-processed datasets to segment words into sub-words by BPE \citep{sennrich-etal-2016-neural}. This helps create a shared vocabulary for source and target languages of all models and avoid out-of-vocabulary entries. Our BPE setting extracts $30$K unique tokens for each of the source and target sides.

One critical hyper-parameter in our model is $\theta$. Considering the selection criterion in Equation \ref{eq:dp-gt}, a small value of $\theta$ allows the majority of tensors to be transferred and hence leads to a minimal reduction in bandwidth. A very large value is also not plausible as it filters lots of tensors and prevents the client from receiving external knowledge. One solution is to run an exhaustive search to find the best value, which clearly is an expensive process and sometimes impossible in the case of FL. Vacillating between different options to set up $\theta$ requires the engagement of both client and server and could in fact be more costly than simply pulling all tensors. To cope with this and also make our results easily reproducible, we simply consider the median of the differences for $\theta$, meaning we only transfer \textit{half} ($50\%$) of the tensors and ignore the rest. The selection criterion determines which half. In $DP_{g}^{c}$, we consider the active half and in $DP_{l}^{c}$ the less active half is exchanged. Our results show that this simple strategy leads to effective communication without compromising much on quality.

\subsection{Centralized Model Training}
Our baseline results are summarized in Table \ref{tab:base}. The evaluation metric used in all experiments is BLEU \citep{papineni2002bleu} computed by SacreBLEU \citep{post-2018-call}.\footnote{\url{https://github.com/mjpost/sacrebleu}} As expected, models work accurately on in-domain data but perform poorly on other domains, specially if the domain is significantly different from  training samples, e.g. the \textit{PHP} model's BLEU score is \textit{zero} when translating \textit{WMT} test sentences.  
\begin{table}[th]
\begin{center}
\begin{tabular}{l c c c c c}
\toprule
&\textit{WMT} & \textit{OS} & \textit{TED} & \textit{PHP} & \textit{UB} \\\midrule
 \textit{WMT} & \textbf{33.66} & 18.57 & 29.22 & 8.04 & 12.41\\ 
 \textit{OS} & 13.66 & \textbf{23.58} & 24.22 & 7.84 & 13.83\\ 
 \textit{TED} & 12.09 & 13.59 & \textbf{29.32} & 6.67 & 10.15\\ 
 \textit{PHP} & 0.00 & 0.26 & 0.26 & \textbf{34.48} & 0.00\\ 
 \textit{UB} &  0.28 & 0.78 & 0.75 & 2.30 & \textbf{30.15}\\ \bottomrule
\end{tabular}
\caption{\label{tab:base} Baseline results for the De$\rightarrow$En direction. Models are trained for $100$K steps. The first column indicates which training set is used to train the engine and other columns show models' performance on different test sets, e.g. [\textit{OS}][\textit{TED}] = 24.22 indicates the BLEU score of a model trained on the \textit{OS} training set and tested on the \textit{TED} test set. The best BLEU score for each test set is bold-faced.}
\end{center}
\end{table}

In order to remedy the poor quality for out-domain data, we train a central model and adapt it to all other domains with two techniques of \textit{data combination} and \textit{chained fine-tuning}. In \textit{data combination}, we simply concatenate all corpora to create a much larger training set. We initialize the central model with \textit{WMT} parameters and retrain it for extra $50$K steps with the new dataset. In \textit{chained fine-tuning}, we do not combine datasets but instead fine-tune the central model sequentially using each domain's training set for additional $50$K steps ($10$K for each), i.e. we start from the \textit{WMT} model, then sequentially fine-tune it over \textit{UB}, \textit{OS}, and other datasets one after another. This sort of fine-tuning could suffer from catastrophic forgetting, so we ran different experiments to figure out the best order of fine-tuning.  

The \textit{chained fine-tuning} strategy could provide relevant baselines for FL experiments. Imagine a scenario where the central model is shipped to a client environment and it is updated there with multiple local datasets. In \textit{data combination}, we assume that all data is accessible at training time (a fully-observable environment with full access to all domains' data) whereas \textit{chained fine-tuning} pictures a more realistic scenario by forcing to update the central model gradually on the client side. 

Table \ref{tab:dc} summarizes results for these two fine-tuning methods. Exposing the central model to other domains' data yields much better quality. \textit{Data combination} clearly outperforms and it shows the impact of having direct access to data; the privilege that we do not have in settings such as \textit{chained training} and FL. 
\begin{table}[th]
\begin{center}
\begin{tabular}{c p{6.5mm} p{6.5mm} p{6.5mm} p{6.5mm} p{6.5mm}}
\toprule
Technique & \textit{WMT} & \textit{OS} & \textit{TED} & \textit{PHP} & \textit{UB}\\\midrule
\textit{chained}  & 18.26 & \textbf{23.51} & 28.19 & 16.14 & 23.05 \\ 
\textit{combination} & \textbf{33.50}& 21.82 & \textbf{31.51} &  \textbf{37.56} & \textbf{35.61} \\ \bottomrule
\end{tabular}
\caption{\label{tab:dc} Domain adaptation results in centralized settings.}
\end{center}
\end{table}
\subsection{Federated Learning Results}
Models reported in the previous section (specially in Table \ref{tab:dc}) are high-quality engines that are trained in a centralized fashion and provide acceptable performance for all domains. Fine-tuning addressed the problem of poor quality for out-domain data, but as  discussed previously, centralized fine-tuning and having access to out-domain data might not be always possible. Therefor, in this section we try to train comparable alternatives in an FL setting. 

Our setting has one server and five clients (one for each dataset). In the interest of fair comparison between the FL and centralized approaches, we initialize all the clients with \textit{WMT} parameters. Each client updates its model with local data and shares it with the server in each round. We {\fontfamily{pcr}\selectfont Pull} client updates after $10$K steps of fine-tuning for aggregation and repeat this process for $5$ rounds. In total, each model is fine-tuned for $50$K steps which is identical to the setting we used for centralized training. Results for this experiment are summarized in Table \ref{tab:fl}.
\begin{table}[th]
\begin{center}
\begin{tabular}{c p{6.5mm} p{6.5mm} p{6.5mm} p{6.5mm} p{6.5mm}}
\toprule
& \textit{WMT} & \textit{OS} & \textit{TED} & \textit{PHP} & \textit{UB} \\\midrule
$\mathcal{S}$ & \textbf{33.97} & 19.17 & 30.8 & 37.32 & 47.9 \\
 \textit{WMT}$^c$ & {32.07} & 18.28 & 29.55 & 9.55 & 13.75 \\
 \textit{OS}$^c$ & 19.05 & \textbf{23.39} & 27.85 & 13.57 & 18.58 \\
 \textit{TED}$^c$ & 17.37 & 16.05 & \textbf{34.30} & 11.83 & 17.05 \\
 \textit{PHP}$^c$ & 4.07 & 4.33 & 7.48 & \textbf{45.07} & 10.90 \\
 \textit{UB}$^c$ & 0.77 & 4.27 & 5.66 & 14.98 & \textbf{49.51} \\ \bottomrule
\end{tabular}
\caption{\label{tab:fl} FL results with $1$ server ($\mathcal{S}$) and $5$ clients (indicated with the $c$ superscript), e.g. \textit{OS}$^c$ is a client initialized with WMT parameters and updated with its own data (\textit{OS} training samples).}
\end{center}
\end{table}

The first row in Table \ref{tab:fl} belongs to the server. FL affects model training quite positively and provides significantly better BLEU scores, specially for those low-performing models such as \textit{PHP}. The average BLEU score of the server over different domains is $33.83$, which is \textbf{1.83} points higher than that of the best model reported in Table \ref{tab:dc}. This means, even though FL does not access clients' data, it is more impactful in fusing information and training mixed-domain engines. This outcome for a complex task such as NMT was unexpected. 

After the final FL round, the server parameters are pushed back to clients so they can also benefit from the result of aggregation. At this point, each client \textit{can} decide to run another phase of fine-tuning with local data over the server parameters. This is a trade-off between being domain specific and remaining generic. Results for this process are listed from the second to last rows in Table \ref{tab:fl}, e.g. \textit{PHP}$^{c}$ after the last {\fontfamily{pcr}\selectfont Push} step has access to all server parameters so its BLEU score on in-domain data is $37.32$, similar to that of the server. It is also able to translate other domains with the average BLEU score of $32.96$. However, as the fifth row shows, if it decides to fine-tune the last parameter set it received from the server with its own local data, its BLEU score increases from $37.32$ to $45.07$ on the in-domain test set, but at the same time it loses its generalization over other domains.

\begin{figure}[htp]
    \centering
    \includegraphics[width=0.4\textwidth]{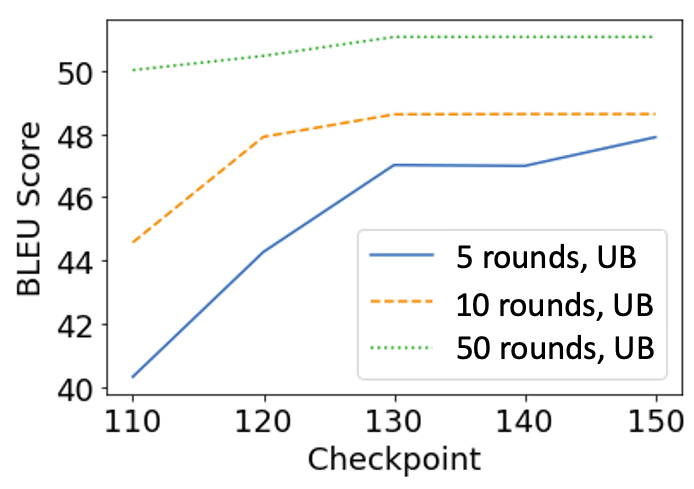}
    \caption{The learning curve of the \textit{UB} model for \textit{rounds} $\in\{5,10,50\}$.}
    \label{fig:num-client}
\end{figure}

We also ran an ablation study to evaluate how the number of FL rounds impact model quality. Two important factors in FL settings are the number of clients and training rounds. We can pass over the first one as we have a cross-silo setting with a limited number of clients, but Table \ref{tab:five-ten} and Figure \ref{fig:num-client} provide additional information on the second hyper-parameter.
\begin{table}[th]
\begin{center}
\begin{tabular}{c p{6.5mm} p{6.5mm} p{6.5mm} p{6.5mm} p{6.5mm}}
\toprule
& \textit{WMT} & \textit{OS} & \textit{TED} & \textit{PHP} & \textit{UB}\\\midrule
$\mathcal{S}_{5}$ & \textbf{33.97} & 19.17 & 30.8 & 37.32 & 47.9\\
$\mathcal{S}_{10}$ & 31.90 & 19.63 & 31.04 & 38.33 & 48.63\\
$\mathcal{S}_{50}$ & 31.05 & \textbf{20.83} & \textbf{32.27} & \textbf{43.99} & \textbf{51.07}\\\bottomrule
\end{tabular}
\caption{\label{tab:five-ten} FL results for \textit{rounds} $\in\{5,10,50\}$. \textit{rounds} = $50$ means aggregation occurs $50$ times between checkpoints $100$K (where FL training starts) and $150$K (where FL training ends).}
\end{center}
\end{table}

Results from the figure/table above show that the number of rounds and model performance increase proportionally in low-quality clients such as \textit{UB} or \textit{PHP}. This was expected since with a higher number of rounds clients are updated more frequently with rich information from the central server. However, it comes at a price as it increases communication load. It also delays local model training, because in each round the client has to suspend training to read server values and updates its internal model. For other high-quality clients such as \textit{WMT}, higher rounds lead to some degradation since they receive external updates from less-accurate peers and have to compensate for their low quality. The choice of the number of rounds is a trade-off between quality and bandwidth.

\subsection{Dynamic Pulling Results}
We proposed a novel technique for better pulling and mentioned that it is able to reduce the communication load yet maintain model quality. Table \ref{tab:low} reports related results to support our claims. 
\begin{table}[th]
\begin{center}
\begin{tabular}{c p{6.5mm} p{6.5mm} p{6.5mm} p{6.5mm} p{6.5mm}}
\toprule
& \textit{WMT} & \textit{OS} & \textit{TED} & \textit{PHP} & \textit{UB}\\\midrule
\textit{default} & 33.97 & 19.17 & 30.8 & 37.32 & 47.90\\\hline
$DP_{l}^{c}$ & 29.28 & \textbf{19.17} & \textbf{30.88} & \textbf{36.33} & \textbf{45.74}\\
$DP_{g}^{c}$ & \textbf{30.74} & 18.28 & 27.61 & 13.69 & 32.16\\
\textit{random} & 24.58 & {19.16} & 30.57 & 35.33 & 42.50\\\bottomrule
\end{tabular}
\caption{\label{tab:low} The impact of different communication techniques on model quality. The first row is copied from Table \ref{tab:fl} for easier comparison. The bold-faced numbers are the best results obtained by DP-based techniques.}
\end{center}
\end{table}

The average BLEU scores for the \textit{default}, $DP_{l}^{c}$, and $DP_{g}^{c}$ methods on different domains are $33.83$, $32.28$, and $24.49$, respectively. This means the assumption that less active tensors are responsible for domain adaption could be true and highly-fluctuating tensors should only be kept for learning in-domain knowledge. We also provide results from our \textit{random} configuration, in which we exchange the same number of parameters as in $DP_{l}^{c}$ and $DP_{g}^{c}$ but those parameters are selected randomly. The comparison between \textit{random} and other alternatives shows that the selection criterion directly affects model quality. 

Although there is a gap by $1.55$ points between \textit{default} and $DP_{l}^{c}$ (which is meaningful in NMT), $DP_{l}^{c}$ could still be a strong candidate when training large models in the context of FL, because the number of parameters exchanged in each pulling step is \textit{45,724,160} for \textit{default} whereas this number is only \textit{22,863,104} ($50$\% less) for $DP_{l}^{c}$. It should also be noted that pulling occurs not once but for multiple rounds and saving $50$\% each time is a significant gain. Moreover, $DP_{l}^{c}$ still performs on par with \textit{data combination} which is a strong but centralized and not private baseline.

\subsection{Comparison to Controllers}
As previously discussed in Section \ref{da}, Controllers \citep{Roosta2021CommunicationEfficientFL} only exchange 4 layers (2 from the encoder and 2 from the decoder in a 12-layer Transformer) with the server in the {\fontfamily{pcr}\selectfont Push} and {\fontfamily{pcr}\selectfont Pull} phases, which means the communication bandwidth they reduce is around $66\%$ ($\frac{4}{12} \approx 33\% $ of the layers are only exchanged between the server and clients). In our case, \textit{DP} only affects {\fontfamily{pcr}\selectfont Pull} which leads to $25$\% bandwidth reduction in client-to-server exchanges.\footnote{{\fontfamily{pcr}\selectfont Pull} is only $50\%$ of the communication where we excluding half the parameters using the threshold, so $50\% \times 50\% = 25\%$.}

We challenged our model to see if we can also save $66\%$ in bandwidth by sending/receiving the same number of parameters as in Controllers. More precisely, we applied our threshold-based strategy in \textit{both} {\fontfamily{pcr}\selectfont Push} and {\fontfamily{pcr}\selectfont Pull} and modified the value of the threshold such that it only accepts $33\% $ of the parameters to exchange with the server. Table \ref{tab:controllers} summarizes results of this experiment. 
\begin{table}[th]
\begin{center}
\begin{tabular}{c p{6.5mm} p{6.5mm} p{6.5mm} p{6.5mm} p{6.5mm}}
\toprule
& \textit{WMT} & \textit{OS} & \textit{TED} & \textit{PHP} & \textit{UB}\\\midrule
DP$^{c}_{l}$ & 29.52 & 19.25 & \textbf{31.53}& \textbf{36.19}&\textbf{45.40}\\
DP$^{c}_{g}$ & 29.09 & 19.07 & 30.93 & 34.47 & 41.57\\
6E6D (0-3) & 31.13	& 19.19& 	30.95&	33.79&	32.85\\
8E8D (0-6)	& \textbf{31.79}	& \textbf{20.02}	&30.6	&32.43	&33.41
\\\bottomrule
\end{tabular}
\caption{\label{tab:controllers} Dynamic parameter selection versus Controllers.}
\end{center}
\end{table}

In our comparison, we selected the two best performing Controller models reported in \citet{Roosta2021CommunicationEfficientFL}. The \textit{6E6D (0-3)} configuration is a Transformer with a 6-layer encoder and 6-layer decoder whose first and fourth layers are selected to act as Controllers. In the \textit{8E8D (0-6)} configuration, instead of using the original encoder/decode layers as Controllers four additional layers (two for the encoder and two for the decoder) are defined which are placed after layers 0 and 6. This means, in an 8-layer encoder/decoder the first and seventh layers are Controllers and the rest are ordinary layers.   

Results show that our model could be a reliable alternative for communication-efficient FL, even though we aggressively limited the number of parameters exchanged in this new setting. Moreover, in our model we do not need to define additional layers. Unlike Controllers, we also do not have to deal with finding the correct position to place Controller layers. According to \citet{Roosta2021CommunicationEfficientFL}, the final model performance is directly impacted by misplacing Controllers and our solution solves that problem.    

\section{Conclusion and Future Work}\label{conc}
In this paper, we reported a set of benchmarking result for NMT in the context of FL. We also proposed an effective technique to reduce the communication bandwidth. Our solution tries to determine a subset of parameters that are responsible for learning out-domain knowledge and only exchanges them with the server. 

In future work, we are interested in \textit{i}) adding more languages to train multilingual engines, \textit{ii}) improving communication protocols even further, \textit{iii}) comparing other FL algorithms in the presence of \textit{DP}, and \textit{iv}) investigating NMT in cross-device settings. 

\section*{Acknowledgments}
We would like to thank our anonymous reviewers for their valuable comments as well as Liling Tan, Grant Strimel, and Sriram Venkatapathy (from Amazon) for providing feedback on the first version of this paper. 

\bibliography{custom}

\begin{thebibliography}{42}
\expandafter\ifx\csname natexlab\endcsname\relax\def\natexlab#1{#1}\fi

\bibitem[{Bahdanau et~al.(2014)Bahdanau, Cho, and Bengio}]{bahdanau2014neural}
Dzmitry Bahdanau, Kyunghyun Cho, and Yoshua Bengio. 2014.
\newblock Neural machine translation by jointly learning to align and
  translate.
\newblock \emph{arXiv preprint arXiv:1409.0473}.

\bibitem[{Bapna et~al.(2019)Bapna, Arivazhagan, and Firat}]{bapna2019simple}
Ankur Bapna, Naveen Arivazhagan, and Orhan Firat. 2019.
\newblock Simple, scalable adaptation for neural machine translation.
\newblock \emph{arXiv preprint arXiv:1909.08478}.

\bibitem[{Bui et~al.(2019)Bui, Malik, Goetz, Liu, Moon, Kumar, and
  Shin}]{bui2019federated}
Duc Bui, Kshitiz Malik, Jack Goetz, Honglei Liu, Seungwhan Moon, Anuj Kumar,
  and Kang~G Shin. 2019.
\newblock Federated user representation learning.
\newblock \emph{arXiv preprint arXiv:1909.12535}.

\bibitem[{Chen et~al.(2019)Chen, Mathews, Ouyang, and
  Beaufays}]{chen2019federated}
Mingqing Chen, Rajiv Mathews, Tom Ouyang, and Fran{\c{c}}oise Beaufays. 2019.
\newblock Federated learning of out-of-vocabulary words.
\newblock \emph{arXiv preprint arXiv:1903.10635}.

\bibitem[{Chu and Wang(2018)}]{chu-wang-2018-survey}
Chenhui Chu and Rui Wang. 2018.
\newblock \href {https://www.aclweb.org/anthology/C18-1111} {A survey of domain
  adaptation for neural machine translation}.
\newblock In \emph{Proceedings of the 27th International Conference on
  Computational Linguistics}, pages 1304--1319, Santa Fe, New Mexico, USA.
  Association for Computational Linguistics.

\bibitem[{Devlin et~al.(2018)Devlin, Chang, Lee, and
  Toutanova}]{devlin2018bert}
Jacob Devlin, Ming-Wei Chang, Kenton Lee, and Kristina Toutanova. 2018.
\newblock Bert: Pre-training of deep bidirectional transformers for language
  understanding.
\newblock \emph{arXiv preprint arXiv:1810.04805}.

\bibitem[{Ge et~al.(2020)Ge, Wu, Wu, Qi, Huang, and Xie}]{ge2020fedner}
Suyu Ge, Fangzhao Wu, Chuhan Wu, Tao Qi, Yongfeng Huang, and Xing Xie. 2020.
\newblock Fedner: Privacy-preserving medical named entity recognition with
  federated learning.
\newblock \emph{arXiv e-prints}, pages arXiv--2003.

\bibitem[{Gehring et~al.(2017)Gehring, Auli, Grangier, Yarats, and
  Dauphin}]{gehring2017convolutional}
Jonas Gehring, Michael Auli, David Grangier, Denis Yarats, and Yann~N Dauphin.
  2017.
\newblock Convolutional sequence to sequence learning.
\newblock In \emph{International Conference on Machine Learning}, pages
  1243--1252. PMLR.

\bibitem[{Geiping et~al.(2020)Geiping, Bauermeister, Dr{\"o}ge, and
  Moeller}]{geiping2020inverting}
Jonas Geiping, Hartmut Bauermeister, Hannah Dr{\"o}ge, and Michael Moeller.
  2020.
\newblock Inverting gradients--how easy is it to break privacy in federated
  learning?
\newblock \emph{arXiv preprint arXiv:2003.14053}.

\bibitem[{Geyer et~al.(2017)Geyer, Klein, and Nabi}]{geyer2017differentially}
Robin~C Geyer, Tassilo Klein, and Moin Nabi. 2017.
\newblock Differentially private federated learning: A client level
  perspective.
\newblock \emph{arXiv preprint arXiv:1712.07557}.

\bibitem[{Hard et~al.(2018)Hard, Rao, Mathews, Beaufays, Augenstein, Eichner,
  Kiddon, and Ramage}]{Hard2018FederatedLF}
Andrew Hard, K.~Rao, Rajiv Mathews, F.~Beaufays, S.~Augenstein, Hubert Eichner,
  Chlo{\'e} Kiddon, and D.~Ramage. 2018.
\newblock Federated learning for mobile keyboard prediction.
\newblock \emph{ArXiv}, abs/1811.03604.

\bibitem[{Hardy et~al.(2019)Hardy, Le~Merrer, and Sericola}]{hardy2019md}
Corentin Hardy, Erwan Le~Merrer, and Bruno Sericola. 2019.
\newblock Md-gan: Multi-discriminator generative adversarial networks for
  distributed datasets.
\newblock In \emph{2019 IEEE international parallel and distributed processing
  symposium (IPDPS)}, pages 866--877. IEEE.

\bibitem[{Hartmann et~al.(2019)Hartmann, Suh, Komarzewski, Smith, and
  Segall}]{hartmann2019federated}
Florian Hartmann, Sunah Suh, Arkadiusz Komarzewski, Tim~D Smith, and Ilana
  Segall. 2019.
\newblock Federated learning for ranking browser history suggestions.
\newblock \emph{arXiv preprint arXiv:1911.11807}.

\bibitem[{Houlsby et~al.(2019)Houlsby, Giurgiu, Jastrzebski, Morrone,
  De~Laroussilhe, Gesmundo, Attariyan, and Gelly}]{houlsby2019parameter}
Neil Houlsby, Andrei Giurgiu, Stanislaw Jastrzebski, Bruna Morrone, Quentin
  De~Laroussilhe, Andrea Gesmundo, Mona Attariyan, and Sylvain Gelly. 2019.
\newblock Parameter-efficient transfer learning for nlp.
\newblock In \emph{International Conference on Machine Learning}, pages
  2790--2799. PMLR.

\bibitem[{Ji et~al.(2020)Ji, Jiang, Walid, and Li}]{ji2020dynamic}
Shaoxiong Ji, Wenqi Jiang, Anwar Walid, and Xue Li. 2020.
\newblock Dynamic sampling and selective masking for communication-efficient
  federated learning.
\newblock \emph{arXiv preprint arXiv:2003.09603}.

\bibitem[{Ji et~al.(2019)Ji, Pan, Long, Li, Jiang, and Huang}]{ji2019learning}
Shaoxiong Ji, Shirui Pan, Guodong Long, Xue Li, Jing Jiang, and Zi~Huang. 2019.
\newblock Learning private neural language modeling with attentive aggregation.
\newblock In \emph{2019 International Joint Conference on Neural Networks
  (IJCNN)}, pages 1--8. IEEE.

\bibitem[{Kim et~al.(2020)Kim, Kim, and Youn}]{kim2020on}
Heejae Kim, Taewoo Kim, and Chan-Hyun Youn. 2020.
\newblock On federated learning of deep networks from non-{\{}iid{\}} data:
  Parameter divergence and the effects of hyperparametric methods.

\bibitem[{Leroy et~al.(2019)Leroy, Coucke, Lavril, Gisselbrecht, and
  Dureau}]{leroy2019federated}
David Leroy, Alice Coucke, Thibaut Lavril, Thibault Gisselbrecht, and Joseph
  Dureau. 2019.
\newblock Federated learning for keyword spotting.
\newblock In \emph{ICASSP 2019-2019 IEEE International Conference on Acoustics,
  Speech and Signal Processing (ICASSP)}, pages 6341--6345. IEEE.

\bibitem[{Li et~al.(2021)Li, Diao, Chen, and He}]{li2021federated}
Qinbin Li, Yiqun Diao, Quan Chen, and Bingsheng He. 2021.
\newblock Federated learning on non-iid data silos: An experimental study.
\newblock \emph{arXiv preprint arXiv:2102.02079}.

\bibitem[{Li et~al.(2019)Li, Wen, Wu, Hu, Wang, Li, Liu, and He}]{li2019survey}
Qinbin Li, Zeyi Wen, Zhaomin Wu, Sixu Hu, Naibo Wang, Yuan Li, Xu~Liu, and
  Bingsheng He. 2019.
\newblock A survey on federated learning systems: vision, hype and reality for
  data privacy and protection.
\newblock \emph{arXiv preprint arXiv:1907.09693}.

\bibitem[{Li et~al.(2020)Li, Sahu, Talwalkar, and Smith}]{9084352}
Tian Li, Anit~Kumar Sahu, Ameet Talwalkar, and Virginia Smith. 2020.
\newblock \href {https://doi.org/10.1109/MSP.2020.2975749} {Federated learning:
  Challenges, methods, and future directions}.
\newblock \emph{IEEE Signal Processing Magazine}, 37(3):50--60.

\bibitem[{Lin et~al.(2018)Lin, Han, Mao, Wang, and Dally}]{lin2017deep}
Yujun Lin, Song Han, Huizi Mao, Yu~Wang, and William~J Dally. 2018.
\newblock Deep gradient compression: Reducing the communication bandwidth for
  distributed training.
\newblock \emph{ICLR}.

\bibitem[{Lison and Tiedemann(2016)}]{lison-tiedemann-2016-opensubtitles2016}
Pierre Lison and J{\"o}rg Tiedemann. 2016.
\newblock {O}pen{S}ubtitles2016: Extracting large parallel corpora from movie
  and {TV} subtitles.
\newblock In \emph{Proceedings of the Tenth International Conference on
  Language Resources and Evaluation ({LREC}'16)}.

\bibitem[{Liu and Miller(2020)}]{liu2020federated}
Dianbo Liu and Tim Miller. 2020.
\newblock Federated pretraining and fine tuning of bert using clinical notes
  from multiple silos.
\newblock \emph{arXiv preprint arXiv:2002.08562}.

\bibitem[{McMahan et~al.(2017)McMahan, Moore, Ramage, Hampson, and
  y~Arcas}]{mcmahan2017communication}
Brendan McMahan, Eider Moore, Daniel Ramage, Seth Hampson, and Blaise~Aguera
  y~Arcas. 2017.
\newblock Communication-efficient learning of deep networks from decentralized
  data.
\newblock In \emph{Artificial Intelligence and Statistics}, pages 1273--1282.
  PMLR.

\bibitem[{Papineni et~al.(2002)Papineni, Roukos, Ward, and
  Zhu}]{papineni2002bleu}
Kishore Papineni, Salim Roukos, Todd Ward, and Wei-Jing Zhu. 2002.
\newblock Bleu: a method for automatic evaluation of machine translation.
\newblock In \emph{Proceedings of the 40th annual meeting of the Association
  for Computational Linguistics}, pages 311--318.

\bibitem[{Passban et~al.(2020)Passban, Saladi, and Liu}]{passban2020revisiting}
Peyman Passban, Puneeth~SM Saladi, and Qun Liu. 2020.
\newblock Revisiting robust neural machine translation: A transformer case
  study.
\newblock \emph{arXiv preprint arXiv:2012.15710}.

\bibitem[{Peters et~al.(2018)Peters, Neumann, Iyyer, Gardner, Clark, Lee, and
  Zettlemoyer}]{Peters:2018}
Matthew~E. Peters, Mark Neumann, Mohit Iyyer, Matt Gardner, Christopher Clark,
  Kenton Lee, and Luke Zettlemoyer. 2018.
\newblock Deep contextualized word representations.
\newblock In \emph{Proc. of NAACL}.

\bibitem[{Pfeiffer et~al.(2020)Pfeiffer, R{\"u}ckl{\'e}, Poth, Kamath,
  Vuli{\'c}, Ruder, Cho, and Gurevych}]{pfeiffer2020adapterhub}
Jonas Pfeiffer, Andreas R{\"u}ckl{\'e}, Clifton Poth, Aishwarya Kamath, Ivan
  Vuli{\'c}, Sebastian Ruder, Kyunghyun Cho, and Iryna Gurevych. 2020.
\newblock Adapterhub: A framework for adapting transformers.
\newblock \emph{arXiv preprint arXiv:2007.07779}.

\bibitem[{Post(2018)}]{post-2018-call}
Matt Post. 2018.
\newblock \href {https://www.aclweb.org/anthology/W18-6319} {A call for clarity
  in reporting {BLEU} scores}.
\newblock In \emph{Proceedings of the Third Conference on Machine Translation:
  Research Papers}, pages 186--191, Belgium, Brussels. Association for
  Computational Linguistics.

\bibitem[{Ren et~al.(2021)Ren, Deng, and Xie}]{Ren2021GRNNGR}
Hanchi Ren, J.~Deng, and Xianghua Xie. 2021.
\newblock Grnn: Generative regression neural network -- a data leakage attack
  for federated learning.

\bibitem[{Roosta et~al.(2021)Roosta, Passban, and
  Chadha}]{Roosta2021CommunicationEfficientFL}
Tanya Roosta, Peyman Passban, and Ankit~R. Chadha. 2021.
\newblock Communication-efficient federated learning for neural machine
  translation.
\newblock \emph{ArXiv}, abs/2112.06135.

\bibitem[{R{\"u}ckl{\'e} et~al.(2020)R{\"u}ckl{\'e}, Geigle, Glockner, Beck,
  Pfeiffer, Reimers, and Gurevych}]{ruckle2020adapterdrop}
Andreas R{\"u}ckl{\'e}, Gregor Geigle, Max Glockner, Tilman Beck, Jonas
  Pfeiffer, Nils Reimers, and Iryna Gurevych. 2020.
\newblock Adapterdrop: On the efficiency of adapters in transformers.
\newblock \emph{arXiv preprint arXiv:2010.11918}.

\bibitem[{Sennrich et~al.(2016)Sennrich, Haddow, and
  Birch}]{sennrich-etal-2016-neural}
Rico Sennrich, Barry Haddow, and Alexandra Birch. 2016.
\newblock \href {https://doi.org/10.18653/v1/P16-1162} {Neural machine
  translation of rare words with subword units}.
\newblock In \emph{Proceedings of the 54th Annual Meeting of the Association
  for Computational Linguistics (Volume 1: Long Papers)}, pages 1715--1725,
  Berlin, Germany. Association for Computational Linguistics.

\bibitem[{Stremmel and Singh(2020)}]{stremmel2020pretraining}
Joel Stremmel and Arjun Singh. 2020.
\newblock Pretraining federated text models for next word prediction.
\newblock \emph{arXiv e-prints}, pages arXiv--2005.

\bibitem[{Sutskever et~al.(2014)Sutskever, Vinyals, and
  Le}]{sutskever2014sequence}
Ilya Sutskever, Oriol Vinyals, and Quoc~V Le. 2014.
\newblock Sequence to sequence learning with neural networks.
\newblock \emph{arXiv preprint arXiv:1409.3215}.

\bibitem[{Tiedemann(2012)}]{TIEDEMANN12.463}
Jörg Tiedemann. 2012.
\newblock Parallel data, tools and interfaces in opus.
\newblock In \emph{Proceedings of the Eight International Conference on
  Language Resources and Evaluation (LREC'12)}, Istanbul, Turkey. European
  Language Resources Association (ELRA).

\bibitem[{Vaswani et~al.(2017)Vaswani, Shazeer, Parmar, Uszkoreit, Jones,
  Gomez, Kaiser, and Polosukhin}]{vaswani2017attention}
Ashish Vaswani, Noam Shazeer, Niki Parmar, Jakob Uszkoreit, Llion Jones,
  Aidan~N Gomez, Lukasz Kaiser, and Illia Polosukhin. 2017.
\newblock Attention is all you need.
\newblock \emph{arXiv preprint arXiv:1706.03762}.

\bibitem[{Wang et~al.(2021)Wang, Lee, Hosseinalipour, Morabito, Chiang, and
  Brinton}]{wang2021device}
Su~Wang, Mengyuan Lee, Seyyedali Hosseinalipour, Roberto Morabito, Mung Chiang,
  and Christopher~G Brinton. 2021.
\newblock Device sampling for heterogeneous federated learning: Theory,
  algorithms, and implementation.
\newblock \emph{arXiv preprint arXiv:2101.00787}.

\bibitem[{Wu et~al.(2020)Wu, Liang, and Wang}]{wu2020fedmed}
Xing Wu, Zhaowang Liang, and Jianjia Wang. 2020.
\newblock Fedmed: A federated learning framework for language modeling.
\newblock \emph{Sensors}, 20(14):4048.

\bibitem[{Yang et~al.(2019)Yang, Liu, Chen, and Tong}]{yang2019federated}
Qiang Yang, Yang Liu, Tianjian Chen, and Yongxin Tong. 2019.
\newblock Federated machine learning: Concept and applications.
\newblock \emph{ACM Transactions on Intelligent Systems and Technology (TIST)},
  10(2):1--19.

\bibitem[{Zhu et~al.(2020)Zhu, Lin, and Zhou}]{zhu2020transfer}
Zhuangdi Zhu, Kaixiang Lin, and Jiayu Zhou. 2020.
\newblock Transfer learning in deep reinforcement learning: A survey.
\newblock \emph{arXiv preprint arXiv:2009.07888}.

\end{thebibliography}
\bibliographystyle{acl_natbib}

\end{document}